\documentclass[runningheads]{llncs}
\usepackage{graphicx}
\usepackage{amsmath}
\usepackage{color}
\usepackage{bbding}

\begin{document}
\title{Relational Subsets Knowledge Distillation for Long-tailed Retinal Diseases Recognition }
\titlerunning{RSKD for Retinal Diseases Recognition}
%
%
\author{Lie Ju\inst{1,2}, Xin Wang\inst{1}, Lin Wang\inst{1, 3}, Tongliang Liu\inst{4}, Xin Zhao\inst{1}, \\ Tom Drummond\inst{2}, Dwarikanath Mahapatra\inst{5} and 
Zongyuan Ge\inst{1,2(}\Envelope\inst{)}}
\authorrunning{L. Ju et al.}
%
\institute{Airdoc LLC, Beijing, China \and
Monash University, Melbourne, Australia \\
\and Harbin Engineering University, Harbin, China \and
 The University of Sydney, Sydney, Australia \and
Inception Institute of Artificial Intelligence, Abu Dhabi, UAE
\\ \url{(https://mmai.group)}
\\
\email{julie@airdoc.com}, \email{zongyuan.ge@monash.edu}}
%

%

%
\maketitle              
\begin{abstract}

In the real world, medical datasets often exhibit a long-tailed data distribution (i.e., a few classes occupy most of the data, while most classes have rarely few samples), which results in a challenging imbalance learning scenario. 
For example, there are estimated more than 40 different kinds of retinal diseases with variable morbidity, however with more than 30+ conditions are very rare from the global patient cohorts, which results in a typical long-tailed learning problem for deep learning-based screening models. 
In this study, we propose class subset learning by dividing the long-tailed data into multiple class subsets according to prior knowledge, such as regions and phenotype information. It enforces the model to focus on learning the subset-specific knowledge. More specifically, there are some relational classes that reside in the fixed retinal regions, or some common pathological features are observed in both the majority and minority conditions. 
With those subsets learnt teacher models, then we are able to distil the multiple teacher models into a unified model with weighted knowledge distillation loss.
The proposed framework proved to be effective for the long-tailed retinal diseases recognition task. The experimental results on two different datasets demonstrate that our method is flexible and can be easily plugged into many other state-of-the-art techniques with significant improvements.



\end{abstract}
\section{Introduction}
\begin{figure}[t]
	\includegraphics[width=10cm]{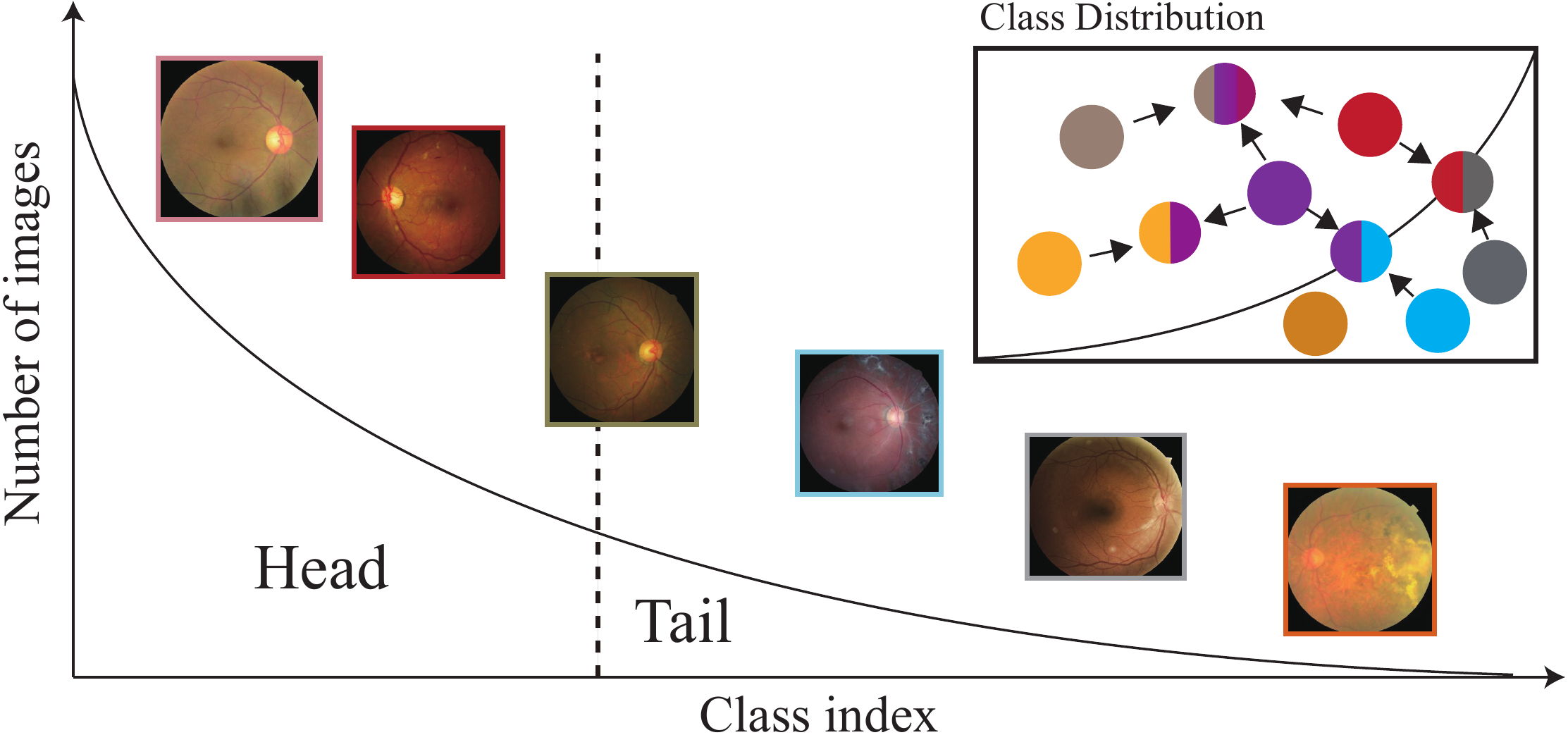}
	\centering
\caption{The retinal disease distribution from~\cite{ju2021improving} exhibits a prominent long-tailed attribute. Label co-occurrence is also a common observation among head, medium and tailed classes (shown in color dots in the class distribution window). } \label{fig_diseases}
\end{figure}

Recent studies have demonstrated successful applications of deep learning-based models for retinal disease screening such as diabetic retinopathy (DR) and glaucoma~\cite{gulshan2016development,fu2018disc}. However, diseases and lesions that occurred less frequently in the training set may not perform well in real clinical test settings due to the algorithm fails to generalize those pathologies.
A recently released dataset~\cite{ju2021improving} collected \textbf{53} different kinds of fundus diseases that may appear in clinical screening. As Fig.~\ref{fig_diseases} shows, this dataset shows a typical long-tailed distribution attribute, whose ratio of majority class (head) to minority class (tail) exceeds 100, indicating a serious class-imbalance issue. This is not uncommon in many medical image datasets.
Besides, some samples may contain more than a single retinal disease label (label co-occurrence), leading to a multi-label challenge.


Most of the existing algorithms to tackle imbalanced long-tailed datasets use either re-sampling~\cite{shen2016relay} or re-weighting~\cite{huang2016learning} strategy. However, since there is label co-occurrence between each class (e.g., macular edema is often accompanied by splenic vein occlusion), the repetitive sampling for those associated instances introduces inner-class imbalance. 
Recently, Wang et al. proposed a multi-task deep learning framework~\cite{wang2019retinal} which can identify 36 kinds of retinal diseases. The data imbalanced is partially solved by bringing all diseases into subsets with regional location information. However, it requires extra annotations for the optic disc and macula's location in the first stage. 
Xiang et al. learned to divide the original samples into multiple class subsets and train several models on those subsets since they often yield better performances than their
jointly-trained counterpart~\cite{xiang2020learning}. However, the multi-label challenge is not discussed and can not be decoupled directly. 

Inspired by the observation that the model trained from less imbalanced class subsets performs better than that trained from the full long-tailed data~\cite{xiang2020learning}, we leverage the prior knowledge of retinal diseases and divide the original long-tailed classes into \textbf{relational} subsets. 
For instance, some diseases (glaucoma, etc.) only occur around the optic disc, while others (epiretinal membrane, macular hole, etc) only affect the macular area, so we can divide those classes into subsets based on pathology region. 
Besides, some diseases may share similar semantics (hypertension, splenic vein occlusion and DR can all cause hemorrhages) in the pathology features. Based on this cognitive law of retinal diseases, we enforce those correlated categories/diseases to be in the same class subset. 
This article proposes three different rules for class subsets partition, which are \textbf{Shot-based}, \textbf{Region-based} and \textbf{Feature-based}. From those subsets, then we train the 'teacher' models accordingly to 'enforce' the model to focus on learning the specific information or features from those class subsets. With all teacher models trained, we later distill the knowledge of several 'teacher' models into one unified 'student' model with an extra weighted knowledge distillation loss, which also contributes to decouple the label co-occurrence in a multi-label setting.


The contributions are summarized as follows: (1) We propose to train the retinal diseases classification model from relational subsets to tackle the long-tailed class distribution problem. 
(2) We leverage the prior knowledge and design three different class partition rules for the relational subsets generation and propose to distill the knowledge from multiple teacher models into a unified model with dynamic weights. 
(3) We conduct comprehensive experiments on two medical datasets \textit{Lesion-10} and \textit{Disease-48} with different tail length. The experimental results demonstrate that our proposed method improves the recognition performance of up to 48 kinds of retinal diseases. Our method can be easily combined with many other state-of-the-art techniques with significant improvements.

\section{Methods}

\begin{figure}[t]
	\includegraphics[width=12cm]{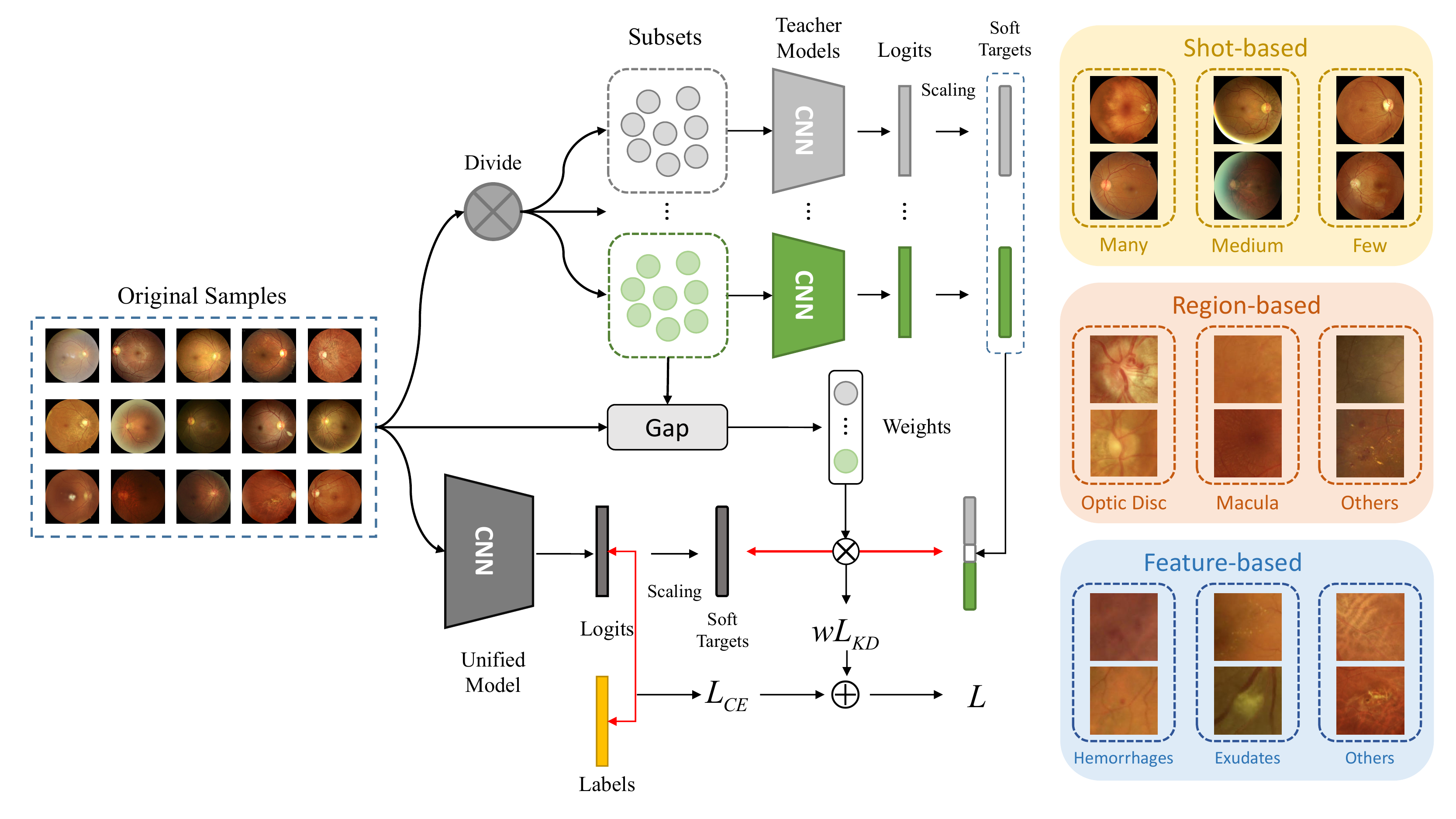}
	\centering
\caption{The overview of our proposed framework. In the first stage, we divide the original long-tailed samples into multiple subsets under the relational class constraint and train separate teacher models. In the second stage, we train a unified student model using a weighted knowledge distillation loss on the trained teacher models' parameters.} \label{fig_framework}
\end{figure}

\begin{figure}[t]
	\includegraphics[width=12cm]{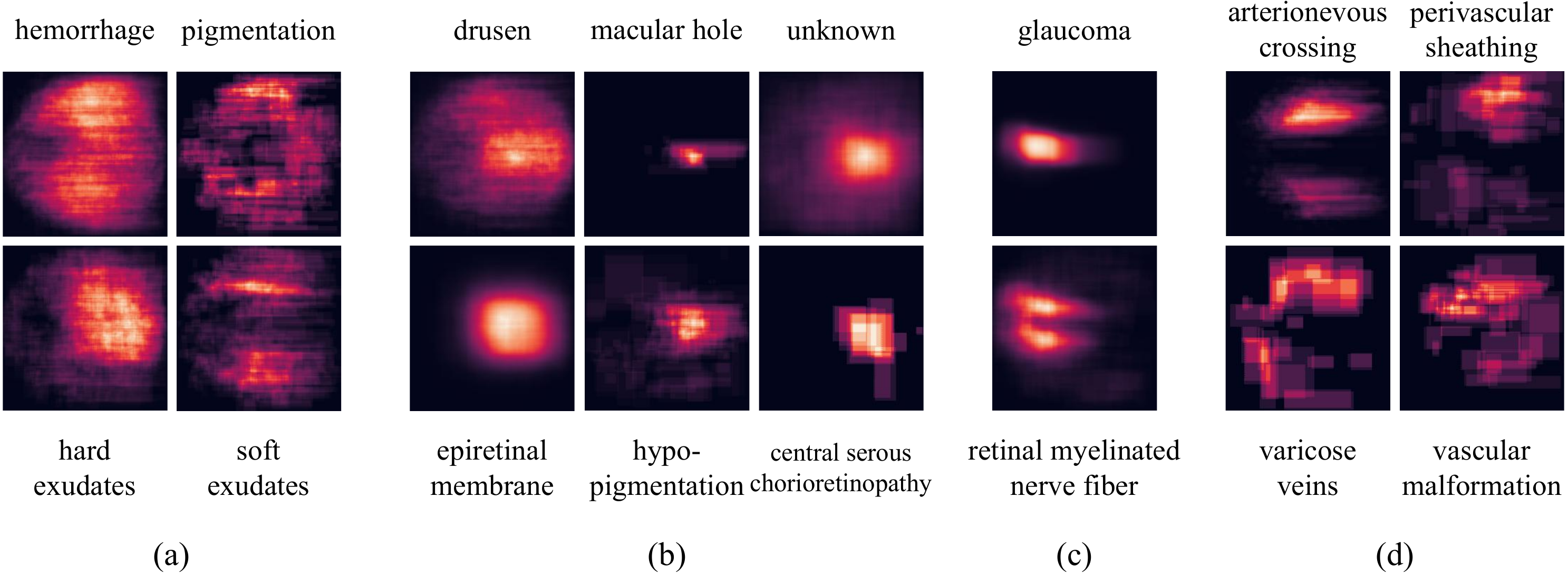}
	\centering
\caption{The heatmap of the locations of various medical signs (lesions) tended to appear frequently in one region: (a) global fundus areas; (b) macula; (c) optic disc; (d) vessels.} \label{fig_pos}
\end{figure}

The overview of our proposed framework is shown in Fig.~\ref{fig_framework}. In the first stage, we divide the original samples into several subsets under the constraint of different rules and train the individual teacher models. In the second stage, we train the unified student model with the knowledge being transferred from the teacher models with a weighted knowledge distillation loss. We show some examples of the regions/lesions in Fig.~\ref{fig_framework} for a better understanding of our methods. 

\begin{table}[t]
\centering
\footnotesize
\label{table_region}
\caption{Region-based subsets and corresponding lesions/diseases.}
\begin{tabular}{c|l}
\hline
\multicolumn{1}{c|}{Region} & \multicolumn{1}{c}{Lesions/Diseases}                                            \\ \hline
Optic disc                  & Glaucoma, retinal myelinated nerve fiber, etc.                                  \\
Macula                      & Epiretinal membrane, drusen, macular hole, hypo-pigmentation, etc.              \\
Vessels                     & Arterionevous crossing phenomenon, perivascular sheathing, etc. \\
Global                      & Hemorrhage, pigmentation, hard exudates, soft exudates,etc.                          \\ \hline
\end{tabular}
\end{table}

\subsection{Relational Subsets Generation}
Formally, given the original long-tailed samples $S_{original}$, we divide them into several subsets \{$S^{1}_{[i_{1},...,i_{o-1}]}$, $S^{2}_{[i_{o},...i_{p}]}$, ..., $S^{k}_{[i_{p+1},...,i_{n}]}$\}, where k denotes the subset ID and $i_{o}$ denotes the $o_{th}$ class ID in the sorted all $n$ classes.  
There are two main advantages: \textbf{(1)} subsets can help reduce the label co-occurrence in a multi-label setting.
More specifically, for $S_{original}$ with $N_{all}$ instances. We randomly select some instances, which belong to class $i$ or $j$. So we have the sampling probability $p_{i} = \frac{N_{i}}{N_{all}}$, $p_{j} = \frac{N_{j}}{N_{all}}$, and $p_{i} \gg p_{j}$. Less label co-occurrence helps to have $p_{j} \rightarrow p_{i}$ as close as possible.
\textbf{(2)} learning from class subsets reduces the length of tail classes and makes the class distribution towards a more balanced state. 
We thus have proposed the following three rules for class subsets generation:

\noindent\textbf{Shot-based.} Inspired by \cite{xiang2020learning}, we divide the long-tailed samples into class-subsets according to their sampling probability, which is equivalent to the number of shots: many, medium and few. Hence, those subsets are all in a relatively less-imbalanced state than that of the original sample distribution, and the learning process involves less interference between the majority and the minority.

\noindent\textbf{Region-based.} Here, we divide the lesions/diseases which are frequently appeared in the specific regions into the same subsets, as Table~\ref{table_region} shows. We leverage this kind of prior knowledge to have the model enforced to focus on the local area in an unsupervised manner. We extracted the locations of lesions in \textit{Lesion-10} dataset and drew heatmaps for various categories to verify our hypothesis, which is shown in Fig.~\ref{fig_pos}.

\noindent\textbf{Feature-based.} Many diseases share similar pathological features. For example, hypertension and DR both manifest phenotypes such as fundus hemorrhage. Therefore, we divide the diseases which share similar semantics into the same class-subsets, and those common features can be shared and learned between the majority and minority. This subset divide rule can also help to improve the model to distinguish the fine-grained local features such as drusen and exudates~\cite{ge2015subset}.

\subsection{Multiple Weighted Knowledge Distillation}
We pre-train the teacher models [$M_{s_{1}}$, ..., $M_{s_{k}}$] on generated subsets. Each teacher model tends to learn specific information under less imbalanced cardinality class samples. For example, Coat's disease, congenital retinoschisis, retina vasculitis and uveitis are the main etiological factors in non-traumatic vitreous hemorrhage (a relational subset). Training from such subsets remains the locality preserving characteristic and utilizes the common feature shared between majority and minority class while the within-class distance in the representation space is minimized. 
After all the teacher models being trained, the obtained logits are used for training the unified student model with KD technique.  
Formally, for the subset $S^{j}$ and its $i_{th}$ class, we have the calibrated soft targets $q_{i}$ and $\hat{q}_{i}$ with respect to the logits $z_{i}$ and $\hat{z}_{i}$ from teacher models and unified model:
\begin{equation}
    q_{i} = \frac{\mathrm{exp}(z_{i}/T)}{\sum_{0,1}\mathrm{exp}(z_{i}/T)}, \, \, \hat{q}_{i} = \frac{\mathrm{exp}(\hat{z}_{i}/T)}{\sum_{0,1}\mathrm{exp}(\hat{z}_{i}/T)},
\end{equation}
where $T$ is the hyper-parameter. Then the knowledge distillation loss for $M_{s_{i}}$ is given by:
\begin{equation}
    L_{KD_{i}} = KL(\hat{q}_{i} | q_{i}) = \sum_{0,1}\hat{q}_{i}\cdot\mathrm{ln}\frac{\hat{q}_{i}}{{q}_{i}},
\end{equation}
where $KL(\cdot)$ is the Kullback-Leibler divergence loss.
Since we obtain the KD loss from multiple teacher models $L_{RS-KD} = [L_{KD_{1}}, ..., L_{KD_{n}}]$ for $n$ sorted classes. So we need to find a way to aggregate them for joint optimizing under one unified model. The most intuitive way is to summing up however may lead to unsatisfactory performance. For instance, NPDRI and NPDRIII both belong to the same subset (medium/hemorrhage-related), and the former has more samples for training, but the semantic information of the latter is more obvious and easier to be learnt by the model, which results in the difference in the performance. In this way, since the performance of some tailed classes of the student model is the same as that of the teacher model, it is unfair to distill all subsets at the same rate and weight. Hence, we expect the teacher models to produce positive and dynamic guidance for the unified model. Here, we compute the performance gap between the teacher and student model as the weights for the KD loss:
\begin{equation}
w_{i} = \left\{\begin{matrix}
 1.0& \mathrm{if }\,  \delta Acc_{M^{i}_{S}} \geq  Acc_{M^{i}_{U,c}} \\ 
\frac{Acc_{M^{i}_{S}} - Acc_{M^{i}_{U,c}}}{Acc_{M^{i}_{S}}\cdot(1-\delta )} & \mathrm{if }\,  \delta Acc_{M^{i}_{S}}  <   Acc_{M^{i}_{U,c}}
\end{matrix}\right.
\end{equation}
where $c$ denotes the $c_{th}$ epoch for the $i_{th}$ class when training the unified model and its performance $Acc_{M^{i}_{U,c}}$. $Acc_{M^{i}_{S}}$ is the performance of teacher model which is a fixed value. The $w$ will be updated after every epoch. $\delta$ is set for controlling the KD decaying schedule. The final loss we aim to optimize becomes:

\begin{equation}
    L = L_{BCE} + \sum_{i}w_{i}L_{KD_{i}}
\end{equation}
where $L_{BCE}$ calculates the original outputs of the softmax and its corresponding ground-truth in the unified model with respect to the fed long-tailed samples.

\section{Experiments}
\label{sec-exp}
\subsection{Experimental Settings}
We use ResNet-50~\cite{he2016deep} pre-trained on ImageNet as our backbone with . The input size of the network is 256 $\times$ 256. We apply Adam to optimize the model. The learning rate starts at $1 \times 10^{-4}$ and reduces ten-fold when there is no drop on validation loss till $1 \times 10^{-7}$ with the patience of 5. $\delta$ is set as 0.6. We use the mean average precision (mAP) as the evaluation metric and all the results are evaluated on test set based on 4-fold cross-validation with a total of 100 epochs. All experiments are implemented with Pytorch and 8 $\times$ GTX 1080Ti GPUs.

\subsection{Data Collections and Annotations}

We conduct our experiments on two datasets \textit{Lesion-10} and \textit{Disease-48} with different tail length. The evaluated datasets consist of two parts: re-labeled public dataset and private dataset from hospitals. We follow~\cite{ju2021improving} and select 10 kinds of lesions and 48 kinds of diseases as our target task. In the ophthalmology field, the occurred lesions always determine the diagnostic results of the specific diseases. Both datasets contain multi-label annotations and the statistics are shown in Table~\ref{table_dataset}. For instance, more than 22,000 images have 2+ kinds of diseases in the \textit{Disease-48}. The dataset is divided for training: validating: testing = 7: 1: 2. 
To verify the idea of our proposed region-based subsets training, we also labeled the locations of lesions (bounding box) for \textit{Lesion-10} dataset. 
We will make the test dataset publicly available to the community for future study and benchmark.

\subsection{Quantitative Evaluation}
Table~\ref{table_results} shows the long-tailed classification results on \textit{Lesion-10} and \textit{Disease-48}. We consider the following baseline method families: (a) ERM model; (b) re-sampling~\cite{shen2016relay}; (c) re-weighting; (d) focal loss~\cite{lin2017focal}; (e) other state-of-the-art methods such as OLTR~\cite{liu2019large}. 
Results in the middle section show that our proposed methods have greatly improved the ERM model with at most 3.29\% mAP, and are also competitive or even better than results comparing with other state-of-the-art methods. 
For those two datasets, our method gives a sharp rise in mAP to the tail classes with 15.18\% mAP and 7.78\% mAP, respectively.
We find that the feature-based relational subsets learning claims the best performance on the \textit{Lesion-10} dataset, and the shot-based relational subsets learning outperforms the other two rules on the \textit{Disease-48} dataset that has a longer tail length. 

\begin{table}[]
\centering

\caption{The statistics (number) of multi-label datasets.}
\label{sec_subsets}
\begin{tabular}{c|c|c}
\hline
Multi-label    & Lesion-10 & Disease-48 \\ \hline
1   & 5929      & 27368      \\
2   & 1569      & 29766      \\
3   & 321       & 17102      \\
3+  & 52        & 5574       \\ \hline
Sum & 7871      & 79810      \\ \hline
\end{tabular}
\label{table_dataset}
\end{table}

\begin{table}[t]
\centering
\caption{Long-tailed classification results (mAP) on Lesion-10 and Disease-48.}
\label{table_results}
\begin{tabular}{c|cccc|c|cccc}
\hline
\textbf{Datasets}             & \multicolumn{4}{c|}{Lesion-10} &  & \multicolumn{4}{c}{Disease-48} \\ \hline
\textbf{Methods}              & head   & medium & tail  & total &  & head   & medium & tail  & total \\ \hline
ERM Model                  & 86.84  & 69.13  & 27.01 & 60.99 &  & 62.80   & 46.87  & 18.60 & 42.76 \\
RS~\cite{shen2016relay}                   & 81.42  & 65.97  & 32.99 & 60.13 &  & 55.57  & 37.55  & 23.99 & 39.04 \\
RW                   & 83.42  & 68.10  & 37.74 & 63.00 &  & 63.41  & \textbf{47.76}  & 20.87 & 44.01 \\
Focal Loss~\cite{lin2017focal}           & 86.84  & 68.67  & 39.60 & 65.04 &  & 62.03  & 47.12  & 22.66 & 43.94 \\
OLTR~\cite{liu2019large}                 & 87.55  & \textbf{70.41}  & 26.99 & 61.65 &  & 60.41  & 45.24  & 22.45 & 42.70 \\
LDAM~\cite{cao2019learning}                 & 86.90  & 69.09  & 30.68 & 62.22 &  & 61.00  & 46.90  & 26.73 & 44.88 \\
DB Loss~\cite{wu2020distribution}              & 88.01  & 69.24  & 39.77 & 65.67 &  & 63.19  & 46.61  & 25.99 & 45.26 \\ \hline
Ours (shot-based)    & 86.87  & 67.93  & 34.31 & 63.04 &  & \textbf{64.17}  & 47.02  & 26.38 & 45.85 \\
Ours (region-based)  & 87.22  & 69.35  & 35.61 & 64.06 &  & 64.15  & 46.99  & 25.33 & 45.49 \\
Ours (feature-based) & 85.42  & 65.24  & 42.19 & 64.28 &  & 62.42  & 46.67  & 21.08 & 43.39 \\ \hline
Ours + DB Loss~\cite{wu2020distribution}       & 84.15  & 69.99  & 38.44 & 64.19 &  & 62.94  & 46.97  & \textbf{27.82} & \textbf{45.91} \\
Ours + OLTR~\cite{liu2019large}          & 85.73  & 68.44  & 30.17 & 61.45 &  & 61.78  & 45.22  & 23.98 & 43.66 \\
Ours + LDAM~\cite{cao2019learning}          & 84.88  & 67.04  & 37.56 & 63.16 &  & 61.01  & 47.31  & 27.00 & 45.11 \\
Ours + Focal Loss~\cite{lin2017focal}    & \textbf{86.97}  & 69.75  & \textbf{42.55} & \textbf{66.42} &  & 63.44  & 47.03  & 25.65 & 45.37 \\ \hline
\end{tabular}
\end{table}

As we declare that our framework has the plug-and-play feature for most frameworks, we validate it with the combination of other state-of-the-art methods. As shown in Table~\ref{table_results}, for \textit{Lesion-10}, LDAM~\cite{cao2019learning} and focal loss~\cite{lin2017focal} are further improved by 0.94\% and 1.38\% mAP. However, DB Loss and OLTR drop by 1.48\% and 0.20\%. We further evaluate our methods on the \textit{Disease-48}, and all methods obtain significant improvements, including DB Loss and OLTR.

\subsection{Ablation Study}
\begin{figure}[t]
	\includegraphics[width=9cm]{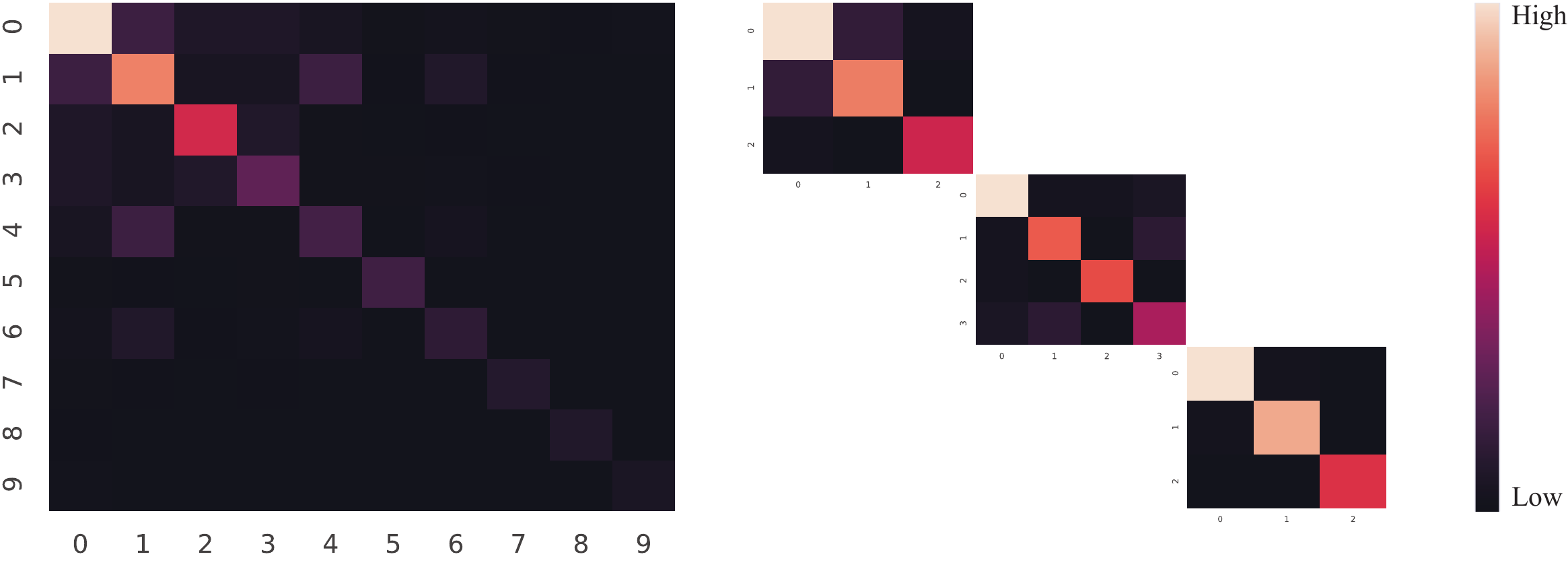}
	\centering
\caption{Left: original distribution. Right: subsets distribution.} \label{fig_decoupling}
\end{figure}



\subsubsection{Visualization of the label co-occurrence.} Table~\ref{table_results} indicates that the regular re-sampling technique does not benefit head and medium classes with performance decrease due to the label co-occurrence. 
As discussed in Sec.~\ref{sec_subsets}, the relational subsets are in a relatively more balanced state by decoupling the original class labels. 
Here, we showcase shot-based subsets learning for the \textit{Lesion-10} dataset in Fig.~\ref{fig_decoupling}. We mainly consider the local co-occurrence labels between classes in the same relational subset with specific semantic information. Then we leverage weighted KD to reduce the risk of regarding the out-of-subsets classes (e.g. classes with label co-occurrence but not included in subsets) as outliers.

\begin{minipage}{\textwidth}

\makeatletter\def\@captype{figure}\makeatother
\begin{minipage}{.4\textwidth}

\centering
\includegraphics[width=5cm]{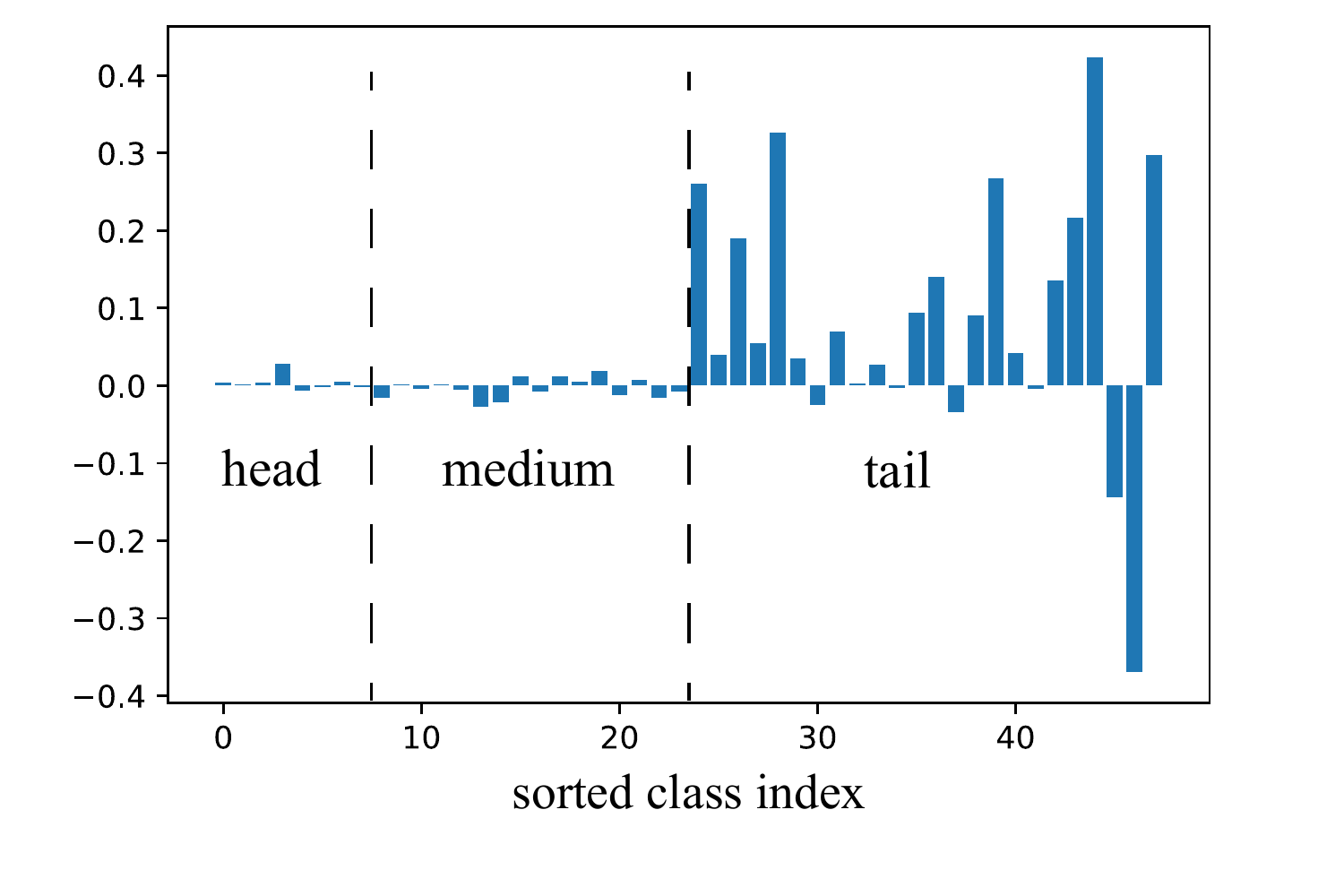}
\caption{The per-class mAP increments between ERM model and ours on Disease-48 dataset.} \label{fig_compare}
\end{minipage}
\makeatletter\def\@captype{table}\makeatother
\begin{minipage}{.55\textwidth}
\centering
\caption{Ablation study results on Lesion-10.}
\label{table_ablation}
\footnotesize
\begin{tabular}{ccccc}
\hline
Methods      & head  & medium & tail  & total \\ \hline
ERM Model    & \textbf{86.84} & \textbf{69.13}  & 27.01 & 60.99 \\
KD (T = 3)   & 86.08 & 67.09  & 32.59 & 61.92 \\
KD (T = 10)  & 86.59 & 66.82  & 36.07 & 63.16 \\
KD (T = 20)  & 85.76 & 68.54  & 30.20  & 61.55 \\
Weighted & 85.42 & 65.24  & \textbf{42.19} & \textbf{64.28} \\ \hline
\end{tabular}
\end{minipage}
\end{minipage}

\subsubsection{The effectiveness of each component.} In order to test how each factor contributes to our model performance, we conduct further ablation studies on various method components. Results are shown in Table~\ref{table_ablation}. Firstly, we assess the sensitivity of knowledge distillation (KD) and how different temperature scaling hyper-parameters affect the model's performance. It can be seen that the KD can only obtain marginal improvements with a low (T = 3) or high temperature (T = 20) value due to the trade-off between CE loss and KD loss. We also show the per-class mAP increments between ERM model and our proposed method on the \textit{Disease-48} dataset in Fig.~\ref{fig_compare}. With the weighted KD Loss, the model has a better detection rate on the tail classes with little performance trade-off on the head and medium classes, which results in an overall gain in performance.

\section{Conclusion}
In this work, we present relational subsets knowledge distillation for long-tailed retinal diseases recognition. We leverage the prior knowledge of retinal diseases such as the regions information, phenotype information and distribution information among all classes, to divide the original long-tailed classes into multiple relational subsets and train teacher models on them respectively to learn a more effective representation. Then we transfer the teacher models into a unified model with the knowledge distillation. Our experiments evaluated on \textit{Lesion-10} and \textit{Disease-48} demonstrate our method can greatly improve the diagnostic result for retinal diseases under the challenging long-tail distribution setting.
\bibliographystyle{splncs04}
\bibliography{z_cite}

\end{document}